\documentclass{article}
\usepackage{spconf,amsmath,graphicx}
\usepackage{multirow}
\usepackage{xspace}
\usepackage{url}
\usepackage{times}
\usepackage{latexsym}
\usepackage{amsmath}
\usepackage{amssymb}
\usepackage{graphicx}
\usepackage{xcolor}

\title{Fast Intent Classification for Spoken Language Understanding}
\name{Akshit Tyagi$^o$, Varun Sharma$^o$, Rahul Gupta$^+$, Lynn Samson$^o$, Nan Zhuang$^o$, Zihang Wang$^o$, Bill Campbell$^+$ } 
\address{$^o$University of Massachusetts, Amherst, USA \\
$^+$Amazon.com, Cambridge, USA}

\begin{document}
\ninept

\maketitle
\begin{abstract}
Spoken Language Understanding (SLU) systems consist of several machine learning components operating together (e.g. intent classification, named entity recognition and resolution).
Deep learning models have obtained state of the art results on several of these tasks, largely attributed to their better modeling capacity.
However, an increase in modeling capacity comes with added costs of higher latency and energy usage, particularly when operating on low complexity devices.
To address the latency and computational complexity issues, we explore a BranchyNet scheme on an intent classification task within SLU systems.
The BranchyNet scheme when applied to a high complexity model, adds exit points at various stages in the model allowing early decision making for a set of queries to the SLU model.
We conduct experiments on the Facebook Semantic Parsing dataset with two candidate model architectures for intent classification.
Our experiments show that the BranchyNet scheme provides gains in terms of computational complexity without compromising model accuracy. We also conduct analytical studies regarding the improvements in the computational cost, distribution of utterances that egress from various exit points and the impact of adding complexity to inference speed and quality.
\end{abstract}

\begin{keywords}
Spoken Language Understanding, BranchyNet, Intent Classification
\end{keywords}

\section{Introduction}
Spoken Language Understanding(SLU) systems are core components of voice agents such as Apple's Siri, Amazon Alexa and Google Assistant and can be designed in one of the several ways, such as an end to end modeling scheme \cite{serdyuk2018towards}, or a collection of task specific classifiers \cite{su2018re,gupta2019one}.
For a complex SLU system, the machine learning architecture can be computationally expensive, posing a challenge for applications such as On-Device-SLU. 
In this work, we explore a scheme that allows us to retain the complexity of the SLU system, while allowing early decision making when possible.
The intuition for such a modeling choice stems for the fact that different queries made to the SLU system may warrant different degrees of processing.
For instance, a request such as ``Stop'' is arguably simpler to process for an SLU system in comparison to a more complex request such as ``find me the closest open restaurant at 8 PM and reserve me a spot''. 
To make use of this specificity in terms of computational load, we use the BranchyNet modeling scheme for an intent classification model as is used in the SLU system described in \cite{su2018re}. 
The BranchyNet scheme allows for decision making at various {\it depths} in a deep learning model, thereby naturally fitting our use case.
The overarching goal of this work is to assess the efficacy of the BranchyNet methodology as a tool to reduce computational complexity and latency, while maintaining model accuracy. 

The SLU system used in \cite{su2018re} uses domain classifiers, intent classifiers, named entity recognizers and reranker in proportion to the number of domains supported.
The domain classification, intent classification and named entity recognizer models identify the domain, user intent and named entities within a request and collectively can take significant compute resources.
Several previous works have suggested methods to reduce computational complexity of the models.
Examples include regularization methods \cite{bishop1995regularization}, model distillation \cite{hinton2015distilling} and compression \cite{cheng2017survey, whittaker2001quantization}.  
All the methods mentioned above attempt to modify the modeling architecture to reduce computational complexity. 
For instance, regularization methods (e.g. L1, dropout) help reduce the number of model parameters and prevent over-fitting. 
However, a further attempt to reduce modeling complexity typically results in an accuracy loss.
Similarly, distillation and compression can prevent over-fitting, but their use for further reduction in computational complexity typically leads to an accuracy loss. 
We consider the BranchyNet\cite{branchynet} as an alternative in this work, as it does not significantly alter the modeling architecture and allows for an adaptive use of fewer parameters depending upon the input query.
The BranchyNet scheme also allows us to fragment the model into multiple sections, allowing for models segments to reside at different locations (e.g. a first few layers/exit points can be stored on device for on-the-edge computing and the rest on the cloud for more complex computation).  

We use the Facebook Semantic Parsing Systems (FSPS) dataset \cite{gupta2018semantic} to evaluate the efficacy of BranchyNet modeling technique.
Using a DNN and stacked-LSTM models as candidate architectures for intent classification, we observe that the introduction of BranchyNet does not lead to any degradation in accuracy.
For the more complex stacked-LSTM model we observe a relative improvement of 10.4\% in the computational complexity (measured in operations per second during inference).
We conduct further analysis on utterances egressing at various exit points in a model and assess the impact of increased model complexity on the egress distribution. 
We provide a more detailed description of the BranchyNet scheme in the next section along with other related work on reducing SLU modeling complexity.

\section{Related work}

\subsection{Fast/Lower complexity SLU}
Previous work that addresses design of fast and lower complexity SLU include use of quantization and hashing \cite{strimel2018statistical}. Another technique uses edge modelling which is private by design \cite{saade2018spoken}. 
Other compression techniques are also used such as knowledge distillation \cite{lu2017knowledge}, matrix factorization \cite{gong2014compressing}, random projection techniques \cite{bingham2001random} and model pruning \cite{han2015deep}.  
These techniques require an initial model choice and often a trade-off is made between accuracy and efficiency.
On the other hand, there exist a set of algorithms that explicitly focus on reducing run time inference complexity. 
Examples include code optimization \cite{vanhouckeimproving} and fast convolution for CNNs \cite{mathieu2013fast, lavin2016fast}.
We experiment with BranchyNet that does not alter the model architecture or inference algorithm. 
This allows us to retain the modeling architecture with the best accuracy, while allowing adaptive early inferences.
We also note that the BranchyNet scheme can be used in combination with any of the aforementioned methods. 

\subsection{BranchyNet scheme}

\label{sec:bn}

\begin{figure}[t]
	\centering
	\includegraphics[trim={0cm 2cm 0 1cm},clip,scale=0.4]{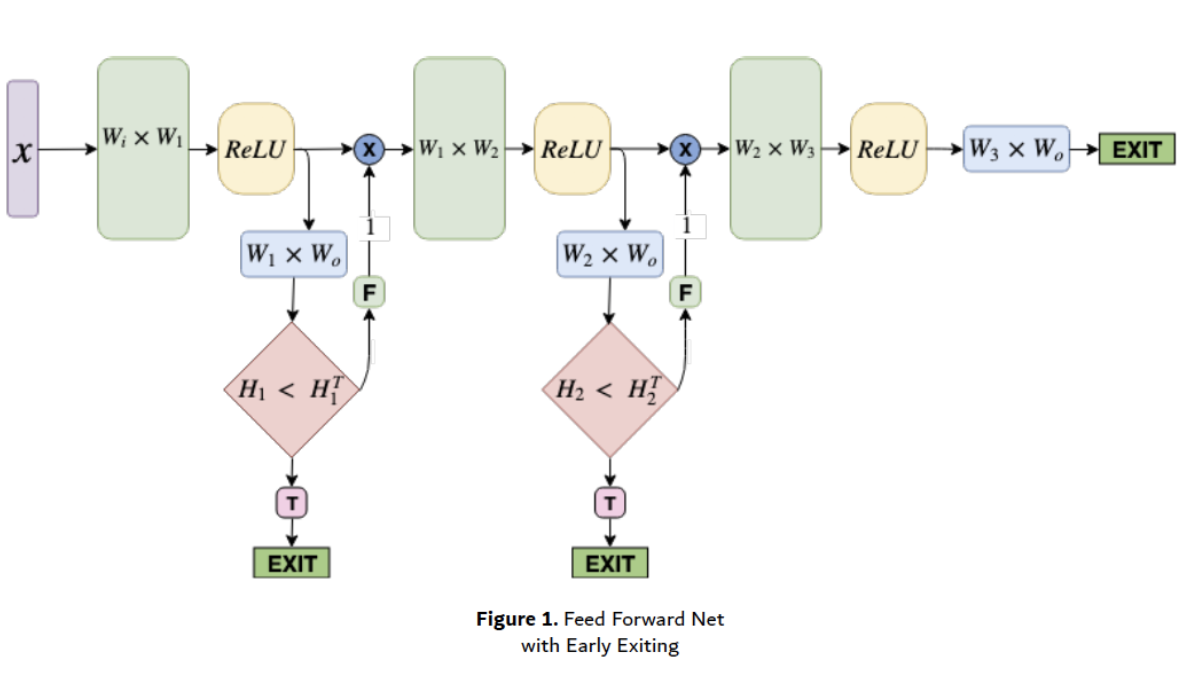}
	\caption{NN Model with early exiting strategy. Exit can be done at each hidden layer based on model confidence.}
  \label{fig:dnn}
\end{figure}

Given a model of choice with multiple potential exit points (e.g. each hidden layer in a DNN), the BranchyNet scheme aims at training a model that makes a decision as soon as it is confident in its prediction.
Given $N$ potential exit points with corresponding feature representations (e.g. outputs obtained at the hidden layer in a DNN), we optimize the following loss function $L$.

\begin{equation}
L = \sum_{n=1}^N \alpha_{n} L_n 
\end{equation}

Where $L_n$ is the cross entropy loss, as determined by the output of the $n^\text{th}$ exit point.
We chose $\alpha_n$ to be a linearly decreasing function of $n$ bounded by values $r_l, r_u$, as shown in Equation~\ref{eq:alpha}.
Such a choice of $\alpha_n$ encourages the learning of discriminative representations in earlier layers, thereby encouraging early exits.
Note that during training, all exit points impact the model parameters and gradients computed using $L_n$ influence the parameters contributing towards feature representation at the exit point $n$.
Apart from learning the exit points for early inference, BranchyNet training also tends to have a regularizing effect on the model parameters as well as prevents vanishing gradients during training. 
 
\begin{equation}\label{eq:alpha}
\alpha_{n} = r_{l} + \frac{r_{u} - r_{l}}{n}, \;\; n = 1,..,N  
\end{equation}

Once trained, an entropy threshold $H_n^T$ is defined for each exit point $n$. 
During inference, the entropy of the output class probabilities is computed at the exit point $n$, starting a scan from the exit point 1 through $N$. 
If the entropy at point $n$ is less than the threshold $H_n^T$, a decision is made at that exit point. 
No further model computations are made for exit points beyond the chosen exit point $n$.
For further details on the BranchyNet model, we refer the reader to \cite{branchynet}.

\begin{figure}[t]
	\centering
	\includegraphics[trim={0cm 15mm 0 0cm},clip,scale=0.4]{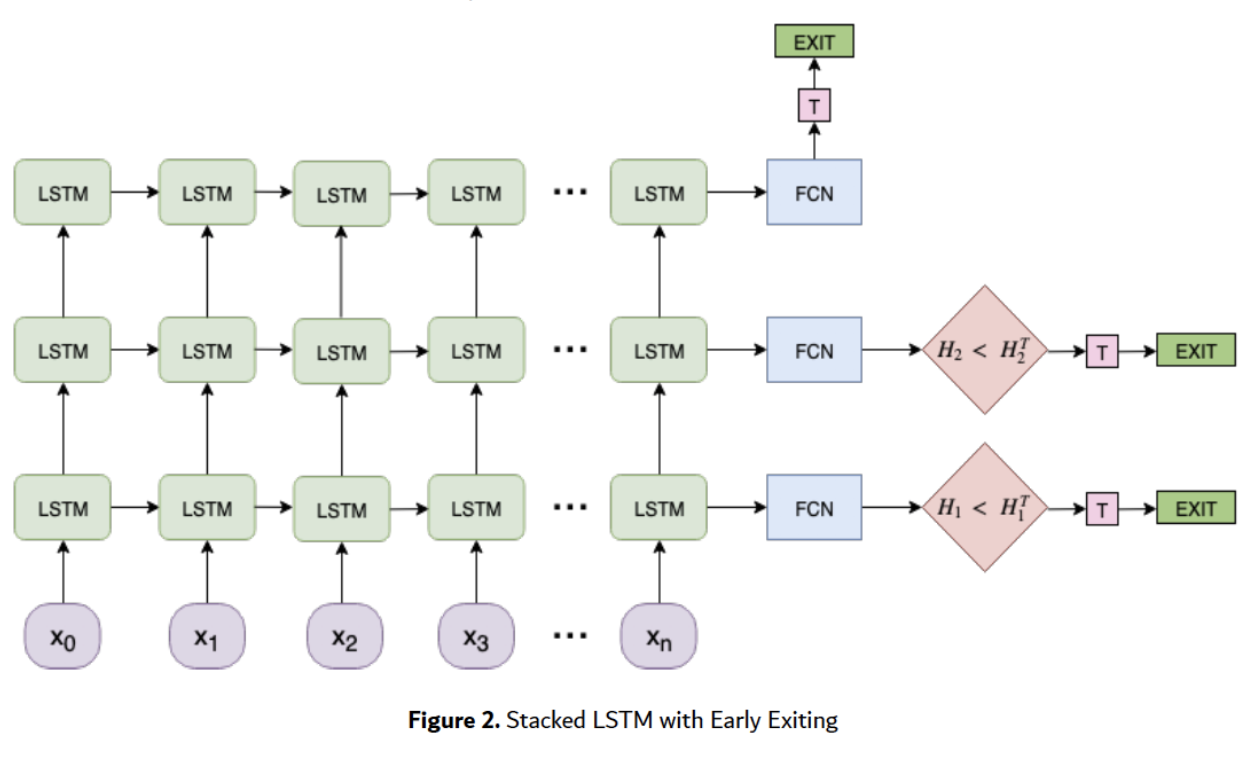}
	\caption{Stacked LSTM model with early exiting strategy. The model can exit at each LSTM layer. FCN implies a fully connected layer.}
  \label{fig:lstm}
\end{figure}

\subsection{Models used in our experimentation}
We use two sets of models for the intent classification task in our experiments, as discussed below.

{\bf Deep Neural Network} We use a DNN with ReLu activations with an exit at each layer. 
The hidden layer outputs from each layer are fed to an output layer as shown in Figure~\ref{fig:dnn}.
The input features to the NN are sentence embeddings computed as mean of the word embeddings constituting the sentence. 
We pre-train the word embedding on a larger Wikipedia corpus \cite{dai2015document} and they are fine tuned during training for the task of our interest. 

{\bf Stacked LSTM} We also apply the BranchyNet technique to a Stacked LSTM network as shown in Figure~\ref{fig:lstm}.
A uni-directional LSTM is applied to the word embeddings and sentence representation is obtained from the hidden layer at the last time step. 
The sentence representation at each LSTM layer is fed to an output layer for decision making.

We train both the above models using the training methodology defined in Section~\ref{sec:bn}.
We tune the number of nodes in the hidden layers in each DNN and Stacked LSTM layer for the best performance on the development set. 
The thresholds $H_n^T$ for $n-th$ exit point are computed as the average value of the entropies obtained on the training data at each of the exit nodes.
At inference time, if any node other than the last exit node obtains an entropy lesser than the average entropy, we make a decision at that exit point.
Else, we make an egress at the last exit point. 
In the next section, we discuss the datasets used in our experimentation.

\section{Dataset}


{\bf Facebook Semantic Parsing System (FSPS) dataset}: The FSPS dataset~\cite{gupta2018semantic} consists of $\sim$44.7k utterances randomly split into $\sim$31.2k training, $\sim$4.4k validation and $\sim$9k test utterances.
The utterances are annotated with a hierarchical representation; however for each given utterance we focus on predicting the overall utterance intent at the top node of the hierarchical annotation.
The dataset consists of 25 intents including get-distance and get-directions intent.
We tune the hyper-parameters of the model (e.g. number of hidden layers, number of stacked LSTM layers etc.) on the development set. 
In particular, we obtain best performance for three layered DNN and stacked-LSTM, leading to three exit points for the respective models.

\section{Results}
We train the DNN and stacked LSTM models on the FSPS datasets with and without the BranchyNet mechanism built into it.
The models without the BranchyNet mechanism serves as our baseline and we compare the change in model performance after adding the mechanism. 
Table~\ref{tab:results} reports the macro F1 score and accuracy on the FSPS dataset.

\begin{table}[t]\label{tab:results}
	\centering
	\begin{tabular}{|l|l|l|} 		\hline
		Model & F1(Macro) & Acc.(\%)    \\ 		\hline
		DNN & 0.48  & 88.5  \\ 		\hline
		DNN + BranchyNet & 0.55  & 89.6 \\		\hline
		Stacked LSTM & 0.65  & 92.8  \\		\hline
		Stacked LSTM + BranchyNet & 0.66  & 93.2  \\		\hline
	\end{tabular}
	\caption{Performance of DNN models on the FSPS dataset with and without the BranchyNet mechanism}
\end{table}

As expected, we observe that the Stacked LSTM outperforms the DNN counterparts.
We also observe a boost in the respective model performances after addition of the BranchyNet mechanism.
While this accuracy can be attributed to the regularization effect (as specified as an added benefit of the BranchyNet scheme in \cite{branchynet}), we also believe that with BranchyNet, each layer can learn to obtain tailored representations for classification at each layer.
In order to quantify the increase in computational complexity, we present analysis in the next section along with an analysis on the utterances egressing from each exit node.
 
\section{Analysis}
We conduct three sets of analysis on the trained BranchyNet network: (i) Evaluating the model size and computational complexity of the trained model, (ii) Understanding the lexical distribution of the utterances exit from each node, (iii) Exit proportions with increased complexity of the models.
  
\subsection{Model size and computational complexity analysis}

\begin{table}[t]
	\centering
	\begin{tabular}{|l|r|r|} 
		\hline
		Exit point & \# Params($\times K$) & FLOPS($\times K$)    \\
    \hline
    \multicolumn{3}{|c|}{3-Layer Neural Network} \\ 
		\hline
		Without BranchyNet & 36.4  & 36.2  \\ 
		\hline
		Exit at first layer & 32.6  & 32.5 \\
		\hline
	  Exit at second layer & 38.9  & 38.7  \\
		\hline
		Exit at final layer & 40.8  & 40.6  \\
		\hline
    \multicolumn{3}{|c|}{Stacked LSTM} \\ 
		\hline
		Without BranchyNet & 22.0 & 69.2  \\ 
		\hline
		Exit at first layer & 7.6 & 23.1 \\
		\hline
	  Exit at second layer & 14.8 & 46.1  \\
		\hline
	  Exit at final layer & 22.1 & 69.2  \\
		\hline
	\end{tabular}
	\caption{FLOPS comparison for performing inference on one datapoint using various models in our experiments}
  \label{tab:flops}
\end{table}

In this section, we compare the cumulative number of parameters before each exit node of a model as well as the number of floating point operations (FLOPS) to arrive at a decision. 
Table~\ref{tab:flops} presents these statistics for the baseline models as well as their versions with the BranchyNet mechanism.  
For the DNN model, we observe that the number of parameters of the overall model (equivalent to the number of parameters while exiting at the final layer) is greater than the baseline models.
This happens due to the fact that each exit point contains parameters for entropy comparison as well as an output layer.
The number of parameters for the Stacked LSTM is lower than three layered DNN due to repeated use of the same set of parameters at each time step.
We also observe that the improvement in computational complexity is higher for Stacked-LSTM. 
For instance, if inference is made at the first layer, it is done with a third of the computational cost when compared to the baseline model without BranchyNet.
Additionally, exit at the final Stacked LSTM layer is almost as expensive as the baseline model, suggesting that the BranchyNet scheme does not add significant computational complexity even in the worst case. 
Overall, we observe that there is an opportunity for reduced computational complexity without taking a hit in the performance, particularly for more complex models.
We look at the distribution of the exit point activation in the next section for the FSPS dataset and report expected reduction in computational complexity depending on this distribution. 

\subsection{Lexical distribution of utterance exit from each node}
\begin{figure}[t]
  \centering
    \includegraphics[trim={1cm 8mm 0 1cm},clip,scale=0.6]{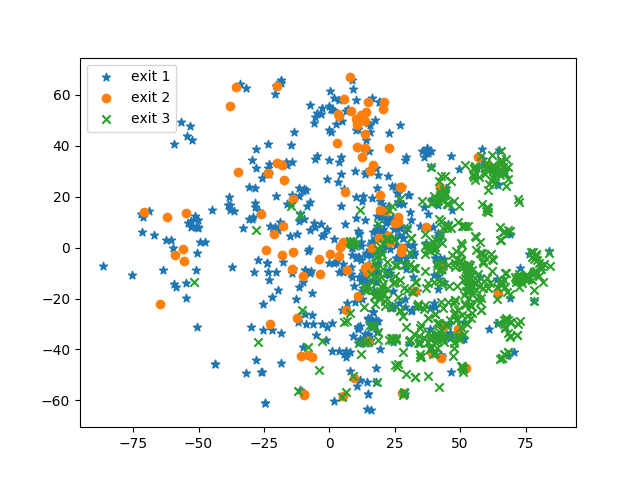}
  \caption{Distribution of utterances as exit from each exit point as obtained after t-SNE projection. Axes are the two dimensions after the t-SNE projection} 
  \label{fig:tsne}
\end{figure}

Given that we observe decrease in computational cost if exiting early, we want to analyse the distribution of exit point activation for the FSPS dataset in Table~\ref{tab:complexity}. 
We observe for the test portion of the FSPS dataset, majority of the test utterances are decoded either in the first layer or the last layer.
Given the exit distribution of the test data, the table also lists the expected number of FLOPs (computed as the sum of FLOPs at each layer, weighted by the probability of exit at that layer).
While we observe an increase in computational complexity of the BranchyNet based DNN model, for the more complex Stacked LSTM model, we observe a relative decrease of 10.4\%. 
This observation advocates for the use of the BranchyNet scheme, particularly when using more complex models.

\begin{table}[t]
	\centering
	\begin{tabular}{|c|c|c|}
    \hline 
		Model & Exit Point & Expected \\
		& activation distribution & complexity \\ \hline
		\multirow{3}{*}{\textbf{DNN}} & {1: 27.80} & \\& {2: 01.92} & \\&  {3: 70.20} & {38.27} \\ \hline
		\multirow{3}{*}{\textbf{Stacked LSTM}} & {1: 15.32} & \\& {2: 00.71} & \\&  {3: 83.97} & {61.99} \\
		\hline
	\end{tabular}
	\caption{Distribution of utterances egressing from each exit point. We also provide the expected complexity based off this observed distribution on the FSPS test set.}
\label{tab:complexity}
\end{table}

In figure~\ref{fig:tsne}, we also plot the distribution of the sentence representations as exit from each exit point.
We extract the sentence embedding of the utterances from a Doc2Vec model pre-trained on a large Wikipedia corpus \cite{dai2015document}.
We train a t-Distributed Stochastic Neighborhood Embedding (t-SNE) model to transform the higher dimensional sentence representations into a 2-dimensional representation for the purposes of visualization.
In the obtained 2-dimensional representation, we observe that the test utterance egressing from each exit point cluster differently in the utterance representation space.
This indicates that there is an inherent difference in the utterances. 
Eyeballing the data egressing from each point reflects the differences in them.
For instance, all the sentences containing the word `restaurant' egress from the first exit point, while all utterances containing the word `holiday' egress from the last exit point.
We also observed that the length of the utterances egressing from the first exit is generally higher than those egressing from other nodes. 
Arguably, the length of the utterance is an indicator of utterance complexity and this observation is against our expectation that more complex utterance will egress from latter exit nodes.
The t-SNE plots along with these observation suggests that each exit node specializes in a specific semantic space and topic as opposed to utterance complexity.

\subsection{Inference sensitivity to Model Complexity}

In this section, we observe the effects of changing the model complexity has on speed of inference and the quality of predictions produced from feed-forward neural network and a Stacked-LSTM model. We add more layers and LSTM stacks respectively, to the models of our interest and observe the exit proportions at each layer.

\begin{figure}[t]
  \centering
    \includegraphics[scale=0.3]{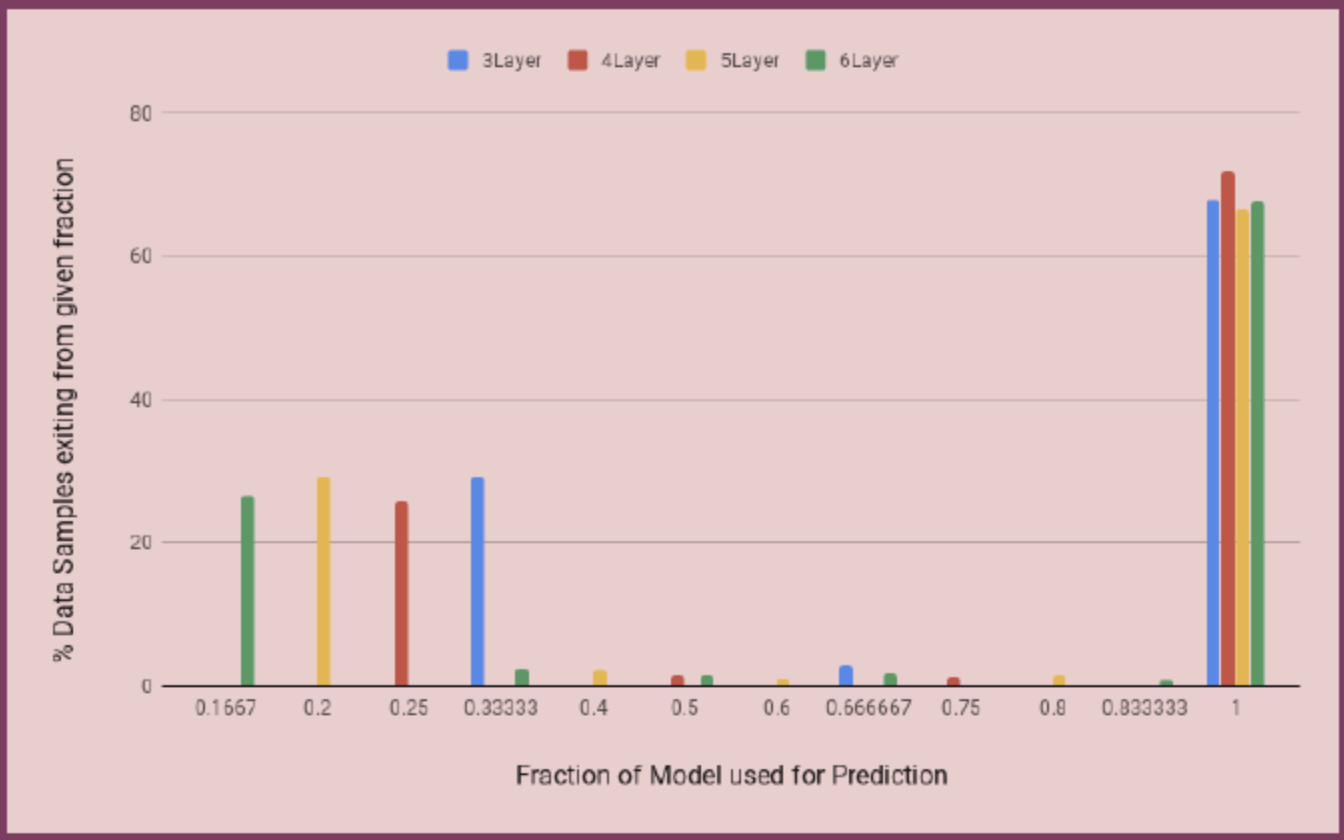}
  \caption{Fraction of test data egressing from each exit point in a 3/4/5/6 layered DNN.}
  \label{fig:dnn_egress}
\end{figure}

\begin{figure}[t]
  \centering
    \includegraphics[scale=0.3]{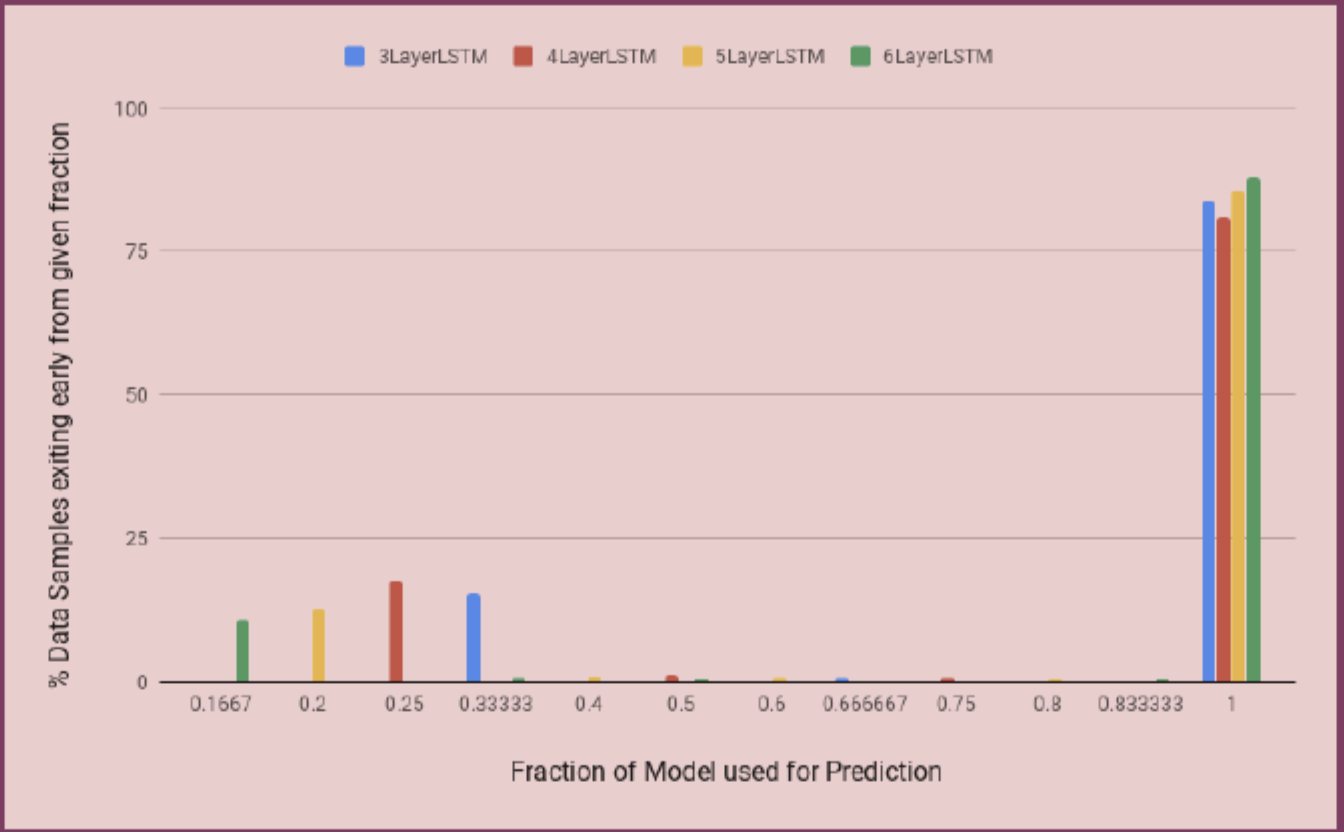}
  \caption{Fraction of test data egressing from each exit point in a 3/4/5/6 layered Stacked-LSTM.}
  \label{fig:slstm_egress}
\end{figure}

Figure~\ref{fig:dnn_egress} and \ref{fig:slstm_egress} presents the exit proportions for the DNN and Stacked-LSTM models.

\subsubsection{Inference Speed}
We use the fraction of layers used to make a prediction (with sufficient confidence) as a substitute for speed, earlier the exit, higher the speed. In both the feed forward neural net and the Stacked LSTM model, we see exit points bunched around the first and the last year and very few model outputs being predicted from the middle layers, hence opening a window for early prediction for a significant chunk of utterances.
\subsubsection{Quality of Model outputs}
We use the t-SNE projection to see the space of utterances and their exit layer number. Figure 3 indicates that the egress proportion of utterances from the first and the last exit point does not change significantly. We hoped that a further break down of the semantic space would be possible with added exit points; providing us flexibility in terms of fragmenting the model (e.g. a variable proportion of the model could be fragmented between device and the cloud). 
However, the results do not indicate that it is possible. We recommend finding the optimal configuration based on expected accuracy maximization as done on the development set in our experiments.

\section{Conclusion}
SLU models often are an ensemble of complex models operating in unison to return machine interpretable hypotheses to downstream components.
Their complexity makes them hard to run on low complexity devices and compression methods often lead to a loss in accuracy.
In this work, we experiment with a BranchyNet scheme that retains the architecture of an intent classification model, yet allowing for reduction in computational complexity. 
The scheme also allows for model fragmentation with a partial model storage on device.
We also perform analysis on the model and observe distribution of utterance egressing from each exit point and the impact of increasing modeling capacity. 
Our results demonstrate the promise of using BranchyNet scheme for SLU systems as a way to reduce computational complexity, without the need to change modeling architecture or take a hit in accuracy.
Our experiments suggest an egress based on semantic clusters and no further splitting of clusters with addition of more exit points.

As the next steps, we aim at extending the BranchyNet scheme to the entire stack of SLU models \cite{su2018re}.
The BranchyNet scheme can also be combined with other modeling schemes such as model distillation and compression.
We also aim to test variants of BranchyNet scheme with exit points added at different points in the model (e.g. a partial set of neurons in a hidden layer) for a further reduction in the complexity.
Exit criterion from each exit point is another parameter for experimentation and we will look for criteria beyond entropy.
Finally, we observed that the egress is based on topic of the utterance. We aim to introduce other forcing functions that encourage egress based on utterance complexity or other factors of interest. 

\bibliography{emnlp2018}
\bibliographystyle{IEEEbib}

\end{document}